\documentclass{article}

\usepackage{arxiv}
\usepackage{multirow} 
\usepackage[utf8]{inputenc} 
\usepackage[T1]{fontenc}    
\usepackage{hyperref}       
\usepackage{url}            
\usepackage{booktabs}       
\usepackage{amsfonts}       
\usepackage{nicefrac}       
\usepackage{microtype}      
\usepackage{lipsum}
\usepackage{graphicx}
\usepackage{threeparttable}
\usepackage{cite}
\usepackage{amsmath,amssymb,amsfonts}
\usepackage{algorithmic}
\usepackage{graphicx}
\usepackage{textcomp}
\usepackage{booktabs}
\usepackage{threeparttable}
\usepackage{tabularx}  
\graphicspath{ {./images/} }

\title{Perturbation Ontology based Graph Attention Networks}

\author{
Yichen Wang$*$ \\
  Institute of Data and Information\\
  Tsinghua University\\
  Shenzhen, China \\
  \texttt{wang-yc22@mails.tsinghua.edu.cn} \\
   \And
 Jie Wang$*$ \\
  College of Eletronic and Information Engineering\\
  Tongji University\\
  Shanghai, China \\
  \texttt{2054310@tongji.edu.cn} \\
  \And
 Fulin Wang \\
  School of Mathematics and Statistics\\
  Shandong University\\
  Shandong, China \\
  \texttt{flwang233@gmail.com} \\
  \And
 Xiang Li \\
  Department of Electrical and Computer Engineering\\
  Carnegie Mellon University\\
  Pittsburgh, USA \\
  \texttt{xl6@andrew.cmu.edu} \\
 \And
 Hao Yin \\
  Institute of Data and Information\\
  Tsinghua University\\
  Shenzhen, China \\
  \texttt{yinh21@mails.tsinghua.edu.cn} \\
 \And
 Bhiksha Raj$\dag$ \\
  Language Technologies Institute\\
  Carnegie Mellon University\\
  Pittsburgh, USA \\
  \texttt{bhiksha@cs.cmu.edu} \\
\thanks{ These authors contributed equally to this work. And $\dag$ corresponding author : bhiksha@cs.cmu.edu.}
}

\begin{document}
\maketitle
\begin{abstract}
In recent years, graph representation learning has undergone a paradigm shift, driven by the emergence and proliferation of graph neural networks (GNNs) and their heterogeneous counterparts. Heterogeneous GNNs have shown remarkable success in extracting low-dimensional embeddings from complex graphs that encompass diverse entity types and relationships. While meta-path-based techniques have long been recognized for their ability to capture semantic affinities among nodes, their dependence on manual specification poses a significant limitation. In contrast, matrix-focused methods accelerate processing by utilizing structural cues but often overlook contextual richness. In this paper, we challenge the current paradigm by introducing ontology as a fundamental semantic primitive within complex graphs. Our goal is to integrate the strengths of both matrix-centric and meta-path-based approaches into a unified framework. We propose perturbation Ontology-based Graph Attention Networks (POGAT), a novel methodology that combines ontology subgraphs with an advanced self-supervised learning paradigm to achieve a deep contextual understanding. The core innovation of POGAT lies in our enhanced homogeneous perturbing scheme designed to generate rigorous negative samples, encouraging the model to explore minimal contextual features more thoroughly. Through extensive empirical evaluations, we demonstrate that POGAT significantly outperforms state-of-the-art baselines, achieving a groundbreaking improvement of up to 10.78\% in F1-score for the critical task of link prediction and 12.01\% in Micro-F1 for the critical task of node classification.
\end{abstract}

\section{Introduction}
\label{sec:intro}

Graphs are a powerful way to represent complex relationships among objects, but their high-dimensional nature requires transformation into lower-dimensional representations through graph representation learning for effective applications. The emergence of graph neural networks (GNNs) has significantly enhanced this process. While early network embedding methods focused on homogeneous graphs, the rise of heterogeneous information networks (HINs) in real-world contexts—like citation, biomedical, and social networks—demands the capture of intricate semantic information due to diverse interconnections among heterogeneous entities. Addressing HIN heterogeneity to maximize semantic capture remains a key challenge.

In HINs, graph representation learning can be classified into two main categories: meta-path-based methods and adjacency matrix-based methods. Meta-path-based approaches leverage meta-paths to identify semantic similarities between target nodes, thereby establishing meta-path-based neighborhoods. A meta-path is a defined sequence in HINs that links two entities through a composite relationship, reflecting a specific type of semantic similarity. For instance, in a social HIN comprising four node types (User, Post, Tag, Location) and three edge types (``interact," ``mark," ``locate"), two notable meta-paths are illustrated: UPU and UPTPU. On the other hand, adjacency matrix-based methods emphasize the structural relationships among nodes, utilizing adjacency matrices to propagate node features and aggregate information from neighboring structures.

Both meta-path-based and adjacency matrix-based methods have notable limitations. Meta-path-based techniques often struggle with selecting effective meta-paths, as the relationships they represent can be complex and implicit. This makes it challenging to identify which paths enhance representation learning, especially in HINs, with diverse node and relation types. The search space for meta-paths becomes vast and exponentially complex, necessitating expert knowledge to identify the most relevant paths. A limited selection can lead to significant information loss, adversely affecting model performance. On the other hand, adjacency matrix-based methods focus on structural information from neighborhoods but often overlook the rich semantics of HINs. While they can be viewed as combinations of 1-hop meta-paths, they lack the robust semantic framework needed to effectively capture implicit semantic information, leading to further information loss.

To address these challenges, we propose using HIN representation learning based on Ontology \cite{giri2011role}, which comprehensively describes entity types and relationships. Ontology models a world of object types, attributes, and relationships \cite{noy2001ontology}, emphasizing its semantic properties. Since HINs are semantic networks constructed based on Ontology, we assert that Ontology provides all necessary semantic information. We define a minimal HIN subgraph that aligns with all possible ontology descriptions as an ontology subgraph. An HIN can be seen as a concatenation of these ontology subgraphs, which offer a complete context for nodes, representing the minimal complete context of each node. Nodes within an ontology subgraph are considered ontology neighbors, forming a local complete context. Compared to meta-paths, ontology subgraphs encompass richer semantics, capturing all node and relation types along with complete context, while meta-paths are limited in scope. Although meta-paths are based on Ontology, ontology subgraphs can capture semantic similarities to some extent. Importantly, the structure of an ontology subgraph is predefined, requiring only a search rather than manual design. In contrast to adjacency matrices, ontology subgraphs represent the smallest complete semantic units with rich semantic information and also provide structural insights due to their natural graph structure. In summary, Ontology combines the strengths of both meta-paths and adjacency matrices.

In this paper, we present Perturbation Ontology-based Graph Attention Networks (POGAT) for graph representation learning that leverages ontology. To improve node context representation, we aggregate both intra-ontology and inter-ontology subgraphs. Our self-supervised training incorporates a perturbation strategy, enhanced by homogeneous node replacement to generate hard negative samples, which helps the model capture more nuanced node features. Experimental results demonstrate that our method surpasses several existing approaches, achieving state-of-the-art performance in both link prediction and node classification tasks.

\begin{table*}[t!]
\begin{center}
\small
\addtolength{\tabcolsep}{-0.6mm}
\caption{Summary of datasets (N types: node types, E types: edge
types, Target: target node, and Classes: Target classes).\label{table.datasets.node}}
\vspace{2mm}

\begin{tabular}{@{}c|rrrrrrr@{}}
\toprule
	 & \# Nodes & \# N Types & \# Edges & \# E Types & Target & \# Classes & \# Task   \\ 
\midrule
DBLP              & 26,128      & 4       & 119,783      & 3         & author            & 4   &LP\&NC\\
IMDB-L               & 21,420     & 4    & 86,642     & 6         & movie            & 4   &NC\\
IMDB-S               & 11,616     & 3    & 17,106     & 2         & -            & -   &LP\\

Freebase      & 43,854   & 4      & 151,034      & 6         & movie            & 3   &NC \\
AMiner         & 55,783      & 3     &  153,676     & 4         & paper            & 4    &LP\&NC\\
Alibaba         & 22,649      & 3     &  45,734     & 5         & -            & -    &LP\\
 \bottomrule
\end{tabular}

\end{center}
\vspace{-2mm}
\end{table*}

\begin{table*}[!t] 
    \centering
    \small
    \addtolength{\tabcolsep}{1pt}
    \caption{Performance evaluation on node classification.}
    \label{tab:node_classification}%

    {\footnotesize In this table, tabular results are in percent; the best result is \textbf{bolded}.}
    \\[2mm] 
    \resizebox{\textwidth}{!}{ 
        \begin{tabular}{@{}l|cc|cc|cc|cc@{}}
        \toprule
        \multirow{2}*{Methods} & \multicolumn{2}{c|}{DBLP} &
        \multicolumn{2}{c|}{IMDB-S} &  \multicolumn{2}{c|}{Freebase} &  \multicolumn{2}{c}{AMiner}\\ 
          & Micro-F1 & Macro-F1  & Micro-F1 & Macro-F1  & Micro-F1 & Macro-F1  & Micro-F1 & Macro-F1   \\\midrule
        \text{GCN}  &91.47 \textpm 0.34 &90.84 \textpm 0.32 &64.82 \textpm 0.64 &57.88 \textpm 1.18 & 68.34 \textpm 1.58 & 59.81 \textpm 3.04 & 85.75 \textpm 0.41 & 75.74 \textpm 1.10\\
        \text{GAT \cite{velivckovic2017graph}}    &93.39 \textpm 0.30 &93.83 \textpm 0.27 & 64.86 \textpm 0.43 &58.94 \textpm 1.35 & 69.04 \textpm 0.58 & 59.28 \textpm 2.56 & 84.92 \textpm 0.68 & 74.32 \textpm 0.95\\ 
        \text{Transformer \cite{vaswani2017attention}} &93.99 \textpm 0.11 &93.48 \textpm 0.12 & 66.29 \textpm 0.69 &62.79 \textpm 0.65 & 67.89 \textpm 0.39 & 63.35 \textpm 0.46 & 85.72 \textpm 0.43 & 74.15 \textpm 0.28\\
        \midrule
        \text{RGCN \cite{schlichtkrull2018modeling}} &92.07 \textpm 0.50 &91.52 \textpm 0.50 &62.95 \textpm 0.15 & 58.85 \textpm 0.26
        & 60.82 \textpm 1.23 &59.08 \textpm 1.44 & 81.58 \textpm 1.44 & 62.53 \textpm 2.31 \\
        \text{HetGNN \cite{zhang2019heterogeneous}} &92.33 \textpm 0.41 & 91.76 \textpm 0.43 &51.16 \textpm 0.65 & 48.25 \textpm 0.67 & 62.99 \textpm 2.31 & 58.44 \textpm 1.99 & 72.34 \textpm 1.42 & 55.42 \textpm 1.45 \\
        \text{HAN \cite{wang2019heterogeneous}} &92.05 \textpm 0.62 &91.67 \textpm 0.49 & 64.63 \textpm 0.58 & 57.74 \textpm 0.96 & 61.42 \textpm 3.56 & 57.05 \textpm 2.06 & 81.90 \textpm 1.51 & 64.67 \textpm 2.21\\
        \text{GTN \cite{yun2019graph}} &93.97 \textpm 0.54 &93.52 \textpm 0.55 &65.14 \textpm 0.45 & 60.47 \textpm 0.98 & - & - & - & -\\
        \text{MAGNN \cite{fu2020magnn}} &93.76 \textpm 0.45 &93.28 \textpm 0.51 &64.67 \textpm 1.67 &56.49 \textpm 3.20 & 64.43 \textpm 0.73 & 58.18 \textpm 3.87 &82.64 \textpm 1.59 & 68.60 \textpm 2.04 \\
        \midrule
        \text{RSHN \cite{zhu2019relation}} &93.81 \textpm 0.55 &93.34 \textpm 0.58 &64.22 \textpm 1.03 & 59.85 \textpm 3.21  & 61.43\textpm5.37 & 57.37 \textpm 1.49 & 73.33 \textpm 2.71 & 51.48 \textpm 4.20  \\
        \text{HetSANN \cite{hong2020attention}} &80.56 \textpm 1.50 &78.55 \textpm 2.42 & 57.68 \textpm 0.44 & 49.47 \textpm 1.21 & - & - & - & -\\
        \text{HGT \cite{hu2020heterogeneous}} &93.49 \textpm 0.25 &93.01 \textpm 0.23 & 67.20 \textpm 0.57 & 63.00 \textpm 1.19 & 66.43 \textpm 1.88 & 60.03 \textpm 2.21 & 85.74 \textpm 1.24 & 74.98 \textpm 1.61 \\
        \text{SimpleHGN \cite{lv2021we}} &94.46 \textpm 0.22 &94.01 \textpm 0.24 &67.36 \textpm 0.57 & 63.53 \textpm 1.36 & 67.49 \textpm 0.97 & 62.49 \textpm 1.69 & 86.44 \textpm 0.48 & 75.73 \textpm 0.97  \\

        \text{HINormer \cite{mao2023hinormer}} &94.94 \textpm 0.21 & 94.57 \textpm 0.23 & 67.83 \textpm 0.34 &64.65 \textpm 0.53  & 69.42 \textpm 0.63 & 63.93 \textpm 0.59 & 88.04 \textpm 0.12 & 79.88 \textpm 0.24\\

        \midrule
        \text{POGAT} &\textbf{96.71} \textpm 0.25 & \textbf{96.21} \textpm 0.22 & \textbf{74.33} \textpm 0.35 &\textbf{72.42} \textpm 0.37  & \textbf{74.12} \textpm 0.49 & \textbf{72.74} \textpm 0.47 & \textbf{93.37} \textpm 0.13 & \textbf{88.24} \textpm 0.28\\
        \bottomrule
        \end{tabular}
    }
    \vspace{-1mm}
\end{table*}

\begin{table}[t]  
\centering 
\vspace{2mm}
\caption{Model performance comparison for the task of link prediction on different datasets.}
\label{table_linkprediction}
\setlength{\tabcolsep}{4pt} 
\small 

\begin{threeparttable}  
    \begin{tabular}{l|ccc|ccc|ccc|ccc}  
    \toprule 
    \multirow{2}{*}{Method} & \multicolumn{3}{c|}{AMiner}  & \multicolumn{3}{c|}{Alibaba} & \multicolumn{3}{c|}{IMDB-L} & \multicolumn{3}{c}{DBLP} \\   
     & R-AUC & PR-AUC & F1 & R-AUC & PR-AUC & F1 & R-AUC & PR-AUC & F1 & R-AUC & PR-AUC & F1 \\   
    \midrule  
    node2vec \cite{grover2016node2vec} & 0.594 & 0.663 & 0.602  & 0.614 & 0.580 & 0.593 & 0.479 & 0.568 & 0.474 & 0.449 & 0.452 & 0.478 \\  
    RandNE \cite{zhang2018billion} & 0.607 & 0.630 & 0.608  & 0.877  & 0.888 & 0.826 & 0.901 & 0.933 & 0.839 & 0.492 & 0.491 & 0.493 \\  
    FastRP \cite{chen2019fast} & 0.620 & 0.634 & 0.600 & 0.927 & 0.900 & 0.926 & 0.869 & 0.893 & 0.811  & 0.515 & 0.528 & 0.506 \\  
    SGC \cite{wu2019simplifying} & 0.589 & 0.585 & 0.567 & 0.686  & 0.708  & 0.623 & 0.826 & 0.889 & 0.769   & 0.601 & 0.606 & 0.587  \\  
    R-GCN \cite{schlichtkrull2018modeling} & 0.599 & 0.601 & 0.610 & 0.674 & 0.710 & 0.629 & 0.826 & 0.878 & 0.790 & 0.589 & 0.592 & 0.566 \\  
    MAGNN \cite{fu2020magnn} & 0.663 & 0.681 & 0.666 & 0.961 & 0.963 & 0.948 & 0.912 & 0.923 & 0.887 & 0.690 & 0.699 & 0.684 \\  
    HPN \cite{ji2021heterogeneous}& 0.658 & 0.664 & 0.660 & 0.958 & 0.961 & 0.950 & 0.900 & 0.903 & 0.892 & 0.692 & 0.710 & 0.687 \\  
    \hline  
    PMNE-n \cite{liu2017principled} & 0.651 & 0.669 & 0.677 & 0.966 & 0.973 & 0.891 & 0.674 & 0.683 & 0.646 & 0.672 & 0.679 & 0.663 \\  
    PMNE-r \cite{liu2017principled}& 0.615 & 0.653 & 0.662 & 0.859 & 0.915 & 0.824 & 0.646 & 0.646 & 0.613 & 0.637 & 0.640 & 0.629 \\  
    PMNE-r \cite{liu2017principled}& 0.613 & 0.635 & 0.657 & 0.597 & 0.591 & 0.664 & 0.651 & 0.634 & 0.630 & 0.622 & 0.625 & 0.609 \\  
    MNE \cite{zhang2018scalable} & 0.660 & 0.672 & 0.681 & 0.944 & 0.946 & 0.901 & 0.688 & 0.701 & 0.681 & 0.657 & 0.660 & 0.635 \\  
    GATNE \cite{cen2019representation} & OOT & OOT & OOT & 0.981 & 0.986 & 0.952 & 0.872 & 0.878 & 0.791 & OOT & OOT & OOT \\  
    DMGI \cite{park2020unsupervised}& OOM & OOM & OOM & 0.857 & 0.781 & 0.784 & 0.926 & 0.935 & 0.873 & 0.610 & 0.615 & 0.601 \\  
    FAME \cite{liu2020fast}& 0.687 & 0.747 & 0.726 & 0.993 & 0.996 & 0.979 & 0.944 & 0.959 & 0.897 & 0.642 & 0.650 & 0.633 \\  
    DualHGNN \cite{xue2021multiplex} & / & / & / & 0.974 & 0.977 & 0.966 & / & / & / & / & / & / \\  
    MHGCN \cite{yu2022multiplex} & 0.711 & 0.753 & 0.730 & 0.997 & 0.997 & 0.992 & 0.967 & 0.966 & 0.959 & 0.718 & 0.722 & 0.703\\  
    BPHGNN \cite{bphgnn} & 0.723 & 0.762 & 0.723 & 0.995 & 0.996 & 0.994 & \textbf{0.969} & 0.965 & 0.943 & 0.726 & 0.734 & 0.731\\  
    \hline 
    POGAT & \textbf{0.804} & \textbf{0.812} & \textbf{0.801} & \textbf{0.998} & \textbf{0.997} & \textbf{0.994} & 0.967 & \textbf{0.986} & \textbf{0.975} & \textbf{0.838} & \textbf{0.819} & \textbf{0.803}\\  
    \hline 
    Std. & \textbf{0.012} & \textbf{0.014} & \textbf{0.011} & \textbf{0.011} & \textbf{0.010} & \textbf{0.011} & \textbf{0.012} & \textbf{0.013} & \textbf{0.012} & \textbf{0.013} & \textbf{0.021} & \textbf{0.012}\\   
    \bottomrule  
    \end{tabular}  
    \begin{tablenotes}  
            \item OOT: Out Of Time (36 hours). OOM: Out Of Memory; DMGI runs out of memory on the entire AMiner data. R-AUC: ROC-AUC.  
    \end{tablenotes}  
\end{threeparttable}

\end{table}

\section{Methods}
\label{sec:format}

With ontology subgraphs as the fundamental semantic building blocks, this section aims to develop a contextual representation of nodes using these subgraphs. Next, we will design training tasks for the network by perturbing the ontology subgraphs.

First of all, we prepare the input node and edge embeddings within an ontology subgraph to be passed to the Graph Transformer Layer (similar to \cite{vaswani2017attention}). For an Ontology sub-graph $\mathcal{G}$ with node features $\alpha_i \in \mathcal{R}^{d_n \times 1}$ for each node $i$ and edge features $\beta_{ij} \in \mathcal{R}^{d_e \times 1}$ for each edge between node $i$ and node $j$, the input node features $\alpha_i$ and edge features $\beta_{ij}$ are passed via a linear projection to embed these to $d$-dimensional hidden features $h_i^{0}$ and $e_{ij}^{0}$.
\begin{equation}
    \label{eqn:input_embd}
    \hat{h}_i^{0} = A^{0} \alpha_i + a^{0} \;\ ; \;\ e_{ij}^{0} = B^{0} \beta_{ij} + b^{0} ,
\end{equation}
where $A^{0} \in \mathcal{R}^{d \times d_n}$, $B^{0} \in \mathcal{R}^{d \times d_e}$ and $a^{0},b^{0}\in \mathcal{R}^{d}$ 
are the parameters of the linear projection layers. We then embed the pre-computed node positional encodings of dim $k$ using a linear projection and add to the node features."$\hat{h}_i^{0}$.
\begin{equation}
\label{eqn:pe_embd_add}
{\lambda}_i^{0} = C^{0} \lambda_i + c^{0} \;\  ; \;\ h_i^{0} = \hat{h}_i^{0} + {\lambda}_i^{0},  
\end{equation}

The Graph Transformer layer closely resembles the transformer architecture originally proposed in \cite{vaswani2017attention}. Next, we will define the node update equations for layer $\ell$.

\begin{equation}
    \label{eqn:gt_layer}
    \hat{h}_{i}^{\ell+1} = O_h^{\ell} \|_{k=1}^{H} \Big(\sum_{j \in \mathcal{N}_i} w_{ij}^{k,\ell} V^{k,\ell}h_j^{\ell} \Big),
\end{equation}
\begin{equation}
    \label{eqn:softmax}
    \textnormal{where,} \ w_{ij}^{k,\ell} = \textnormal{softmax}_j \Big(\frac{Q^{k, \ell} h_i^{\ell} \ \cdot \ K^{k, \ell}h_j^{\ell}}{\sqrt{d_k}}  \Big),
\end{equation}
and $Q^{k,\ell}, K^{k,\ell}, V^{k,\ell} \in \mathcal{R}^{d_k \times d}$, $O_h^{\ell} \in \mathcal{R}^{d \times d}$, $k=1$ to $H$ denotes the number of attention heads, and $\|$ denotes concatenation. 

To ensure numerical stability, the outputs after exponentiating the terms inside the softmax are clamped between $-5$ to $+5$. The attention outputs $\hat{h}_{i}^{\ell+1}$ are then passed to a Feed Forward Network, which is preceded and followed by residual connections and normalization layers, as follows:
\begin{eqnarray}
    \hat{\hat{h}}_{i}^{\ell+1} &=& \textnormal{LayerNorm} \Big( h_{i}^{\ell} + \hat{h}_{i}^{\ell+1} \Big), \label{eqn:rc_norm1}\\
    \hat{\hat{\hat{h}}}_{i}^{\ell+1} &=& W_2^{\ell} \textnormal{ReLU}(W_1^{\ell} \hat{\hat{h}}_{i}^{\ell+1}), \label{eqn:ffn}\\
    h_{i}^{\ell+1} &=& \textnormal{LayerNorm} \Big( \hat{\hat{h}}_{i}^{\ell+1} + \hat{\hat{\hat{h}}}_{i}^{\ell+1} \Big), \label{eqn:rc_norm2}
\end{eqnarray}
where $W_1^{\ell}, \in \mathcal{R}^{2d \times d}$, $W_2^{\ell}, \in \mathcal{R}^{d \times 2d}$, $\hat{\hat{h}}_{i}^{\ell+1}, \hat{\hat{\hat{h}}}_{i}^{\ell+1}$ denote intermediate representations. The bias terms are omitted for clarity.

Given that each ontology subgraph \( \mathcal{O}_i \) 
associated with the target node\( u \)  independently yields an intra-aggregation representation, it becomes imperative to integrate the rich semantic information emanating from each of these subgraphs within the broader network $\mathcal{N}$ via an inter-aggregation process. Considering the minimal context semantic should be equivalent to each other, we turn to use multi-head attention mechanisms to aggregate the semantic information between ontology subgraphs:
  
\begin{equation}
\mathbf{h}_u^{(l)}=\operatorname{ConCat}(\sigma(\mathbf{h}_u^{O(i), k, (l)})),
\end{equation}
where \( k \) is the number of attention heads, $\operatorname{Concat}(\cdot)$ denotes the concatenation of vectors, and we obtain the representation of the last layer by averaging operation:
\begin{equation}
    \mathbf{h}_u^{(L)}=\frac{1}{K}\sum_{k}\mathbf{h}_u^{k, (L)}.
\end{equation}

\subsection{Bi-level perturbation Ontology Training}
To enhance the model's ability to capture the intrinsic semantics of ontology, we employ a perturbation technique to modify the ontology. We also design two specific tasks to differentiate perturbation subgraphs at both the node level and the graph level. 

\subsubsection{Ontology Subgraph perturbation}
\leavevmode \newline
In this section, we enhance the perturbation operation on ontology subgraphs to generate negative samples for self-supervised tasks. Initially, we tried the common all-zero mask, which replaces node embeddings with zero vectors, but this approach yielded unsatisfactory results. Drawing inspiration from \cite{jin2021gcn}, which used random graphs as noise distributions, we then implemented a random mask that selects nodes randomly for substitution, resulting in some improvement. However, given the significant differences in information among various node types, using random nodes can create negative samples that are too dissimilar to the positive samples, making the task easier and potentially reducing model performance. To address this, we further refined our strategy by substituting nodes with similar types, thereby constructing challenging negative samples that enhance the model's ability to learn from minimal contexts.

We take the ontology subgraph set (i.e., \( \mathcal{O}_\text{sub} \)) as positive samples. Then, we randomly replaced nodes in the subgraphs with nodes of the same type to preserve a certain level of semantics similarity. These substitute nodes are marked with diagonal lines. If the generated perturbation subgraph is not included in the original ontology subgraph set, it is labeled as a negative sample and denoted as \( \mathcal{O}_i^m \). The set of all negative ontology subgraphs is denoted as \( \mathcal{O}_\text{sub}^m \). Next, we perform shuffle operations on all positive and negative samples, further readout the context representations of nodes to obtain a graph-level representations of $\mathcal{O}_j$:
\begin{equation}
    \mathbf{h}_G^{\mathcal{O}_j}=\operatorname{ReadOut}(\mathbf{h}_u \mid \forall u \in \mathcal{O}_j, \mathcal{O}_j \in \mathcal{O}_\text{sub}\cup\mathcal{O}_\text{sub}^m)
\end{equation}

\subsubsection{Graph-level Discrimination}
\leavevmode \newline
For graph-level training, we designed a graph discriminator based on an MLP with to determine whether the subgraph has been perturbed:
\begin{equation}
\mathbf{y}_{\text {pred}, G}=\text{Discriminator}_G\left(\mathbf{h}_G^{\mathcal{O}_j}\right) 
\end{equation}

Then we calculate the cross-entropy loss:
\begin{equation}
\mathcal{L}_G=\sum_{O_j} \operatorname{\textsc{CrossEnt} }\left(\mathbf{y}_{\text {pred}, G}, \mathbf{y}_{\text {true}, G}\right),
\end{equation}
where $\mathbf{y}_{\text{true},G}$ stands for the labels of graph-level task.

\subsubsection{Node-level Discrimination}
\leavevmode \newline
Given the node representation $\mathbf{h}_v$ for node $v$, we further employ an MLP $\phi_{\text{MLP}}(\cdot;\theta_{\text{pdt}})$ parameterized by $\theta_{\text{pdt}}$ to predict the class distribution as follows, 

\begin{equation} \label{eq.pred_class}
    \tilde{\mathbf{y}}_{v} = \phi_{\text{MLP}}(\mathbf{h}_v;\theta_{\text{pdt}}).
\end{equation}
where $\tilde{\mathbf{y}}_{v} \in \mathbf{R}^{C}$ is the prediction and $C$ is the number of classes. 
In addition, we further add an $L_2$ normalization on $\tilde{\mathbf{y}}_{v}$ for stable optimization.

Given the training nodes $V_{\text{tr}}$, for multi-class node classification, we employ cross-entropy as the overall loss, as
\begin{equation}    \mathcal{L}_N=\sum_{v\in V_{\text{tr}}}\textsc{CrossEnt}(\tilde{\mathbf{y}}_{v},\mathbf{y}_v),
\end{equation}
where $\textsc{CrossEnt}(\cdot)$ is the cross-entropy loss, and $\mathbf{y}_v\in\mathbf{R}^C$ is the one-hot vector that encodes the label of node $v$.
Note that, for multi-label node classification, we can employ binary cross-entropy to calculate the overall loss.

Finally, we performed joint training on both tasks, allowing our model to learn minimal context semantics from both graph-level and node-level perspectives. We optimized the model by minimizing the final objective function:
\begin{equation}
   \mathcal{L}=\gamma \cdot \mathcal{L}_N+(1-\gamma)\cdot \mathcal{L}_G,
\end{equation}
where $\gamma \in \left[ 0, 1\right]$ is a balance scalar.

\section{Experiments}
In this section, we perform a comprehensive set of experiments to assess the effectiveness of our proposed method, POGAT, specifically targeting node classification and link prediction tasks. Our goal is to showcase the superiority of POGAT by comparing its performance with existing state-of-the-art methods.

\subsection{Datasets.} 

Our experimental evaluation spans across six publicly available, real-world datasets: IMDB-L (dataset1), IMDB-S (dataset2), Alibaba (dataset3), DBLP (dataset4), Freebase (dataset5), and Aminer (dataset6). A concise summary of each dataset's statistical properties is provided in Table 1. For all baselines, we use their released source code and the parameters recommended by their papers to ensure that their methods
achieve the desired effect.

\subsection{Node classification.} 

We conduct a comprehensive evaluation of our model's efficacy in node classification tasks by comparing it against state-of-the-art baselines. The results of this evaluation are detailed in Table 2, where the best scores are highlighted in bold for clarity and emphasis. Our proposed POGAT model demonstrates a remarkable performance advantage, significantly surpassing all baseline models in both Macro-F1 and Micro-F1 metrics across a diverse range of heterogeneous networks. This robust performance indicates the effectiveness of our approach in capturing the underlying structures and relationships within the data. For DBLP and IMDB-S, we leverage standard settings and benchmark against the HGB leaderboard results. For the remaining datasets, we adhere strictly to the default hyperparameter settings of the baseline models. Furthermore, we fine-tune these hyperparameters based on validation performance to optimize the results.

\subsection{Link prediction.} 

Next, we evaluate POGAT's performance in unsupervised link prediction against leading baselines. The results of this evaluation are comprehensively summarized in Table 3, which provides a clear illustration of the model's effectiveness across various tested networks. Our findings reveal that POGAT achieves state-of-the-art metrics in link prediction, showcasing its capability to effectively identify and predict connections within complex network structures. Notably, POGAT demonstrates an average improvement of 5.92\%, 5.42\% and 5.54\% in R-AUC, PR-AUC, and F1, respectively, over the GNN MHGCN on six datasets.

\section{Conclusion}
\label{sec:refs}
In conclusion, this research addresses the challenges of heterogeneous network embedding through the introduction of Ontology. We present perturbation Ontology-based Graph Attention Networks, a novel approach that integrates ontology subgraphs with an advanced self-supervised learning framework to achieve a deeper contextual understanding. Experimental results on six real-world heterogeneous networks demonstrate the effectiveness of POGAT, showcasing its superiority in both node classification and link prediction tasks.


\end{document}